\documentclass[letterpaper, 10 pt, conference]{ieeeconf}
\IEEEoverridecommandlockouts
\usepackage{graphicx}
\usepackage{xcolor}
\usepackage{amsmath}
\usepackage{hyperref}
\usepackage{url}
\usepackage{mathtools}
\usepackage{commath}
\usepackage[ruled,vlined]{algorithm2e}
\usepackage{siunitx}
\usepackage{xcolor}

\title{\LARGE \bf
GarNet: A Continuous Robot Vision Approach for Predicting Shapes and Visually Perceived Weights of Garments
}

\author{Li Duan$^{1}$ and Gerardo Aragon-Camarasa$^{1}$
\thanks{$^{1}$ Li Duan and Gerardo Aragon-Camarasa are with the School of Computing Science, University of Glasgow, G12 8QQ, Scotland, United Kingdom {\tt\small l.duan.1@research.gla.ac.uk}  and {\tt\small gerardo.aragoncamarasa@glasgow.ac.uk}}%
}
\begin{document}
\maketitle

\begin{abstract}
We present a Garment Similarity Network (GarNet) that learns geometric and physical similarities between known garments by continuously observing a garment while a robot picks it up from a table. The aim is to capture and encode geometric and physical characteristics of a garment into a manifold where a decision can be carried out, such as predicting the garment's shape class and its visually perceived weight. Our approach features an early stop strategy, which means that GarNet does not need to observe a garment being picked up from a crumpled to a hanging state to make a prediction. In our experiments, we find that GarNet achieves prediction accuracies of $92\%$ for shape classification and $95.5\%$ for predicting weights and advances state-of-art approaches by $~21\%$ for shape classification.
\end{abstract}

\section{Introduction\label{sec:intro}}

Deformable objects remains an open problem for robotic perception and manipulation. This is because garments have an infinite number of possible configurations, such as crumpled and irregular shapes, that cannot be modelled easily via simulations \cite{wang2011data,hoque2021visuospatial}. Specifically, the main challenges in deformable objects perception and manipulation are twofold. First, they usually have a complex initial configuration, which means they are wrinkled, crumpled or folded, and not in a known configuration state that can be used for manipulation tasks. Second, garments usually deform in unpredictable ways, making predictions of their deformations difficult during dexterous robotic manipulations.

Robots manipulating deformable objects without prior knowledge about their geometric and physical properties (e.g. shapes, weights or stiffness parameters) can result in robots requiring to plan actions using a complex and high-dimensional space. This, therefore, causes failures in motion planning since robots are prone to fail due to minor variations in the deformable object's configurations. We, therefore, propose in this paper the first step towards an online continuous perception approach that equips a robot with the ability to predict garment shapes and, consequently, allows a robot to stop a grasping or manipulation task if the prediction belief is above a threshold.

\begin{figure}[t]
    \centering
    \includegraphics[width=0.5\textwidth]{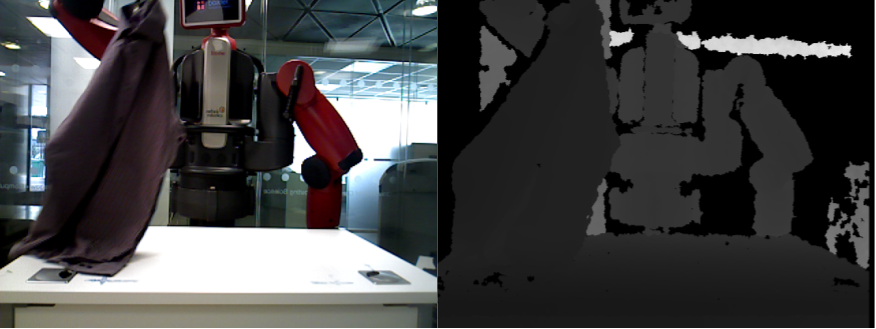}
    \caption{An example of our experimental setup for capturing a database of garment's deformations. That is, garments are grasped and dropped by a dual-armed robot. An Xtion camera is placed in the front of the robot to record a video sequence, consisting of RGB images and depth images.
    }
    \label{fig:garment_manipulation}
\end{figure}

Compared to traditional research focusing on wrinkles \cite{sun2016recognising} and other local features~\cite{sun2017singleshot,MARTINEZ2019220}, we propose to learn the dynamic properties of garments from video sequences and allow a robotic system to recognise the shape and weight of a garment continuously. For this, we devise a Garment similarity Network (GarNet) that learns the physical similarity between garments to predict shapes (geometric) and visually perceived weights (physical) of unseen garments. We define a visually perceived weight in three discretised levels using an electric scale to physically weigh garments; namely, light, medium and heavy weights. We hypothesise \textit{that GarNet can predict shapes and discretised weights in approximately 0.1 seconds per image frame by learning geometric and physical properties and predicting the garment's shape and weights continuously during a robotic garment pick and place task.}

\begin{figure*}[t]
    \centering
    \includegraphics[width=0.95\textwidth]{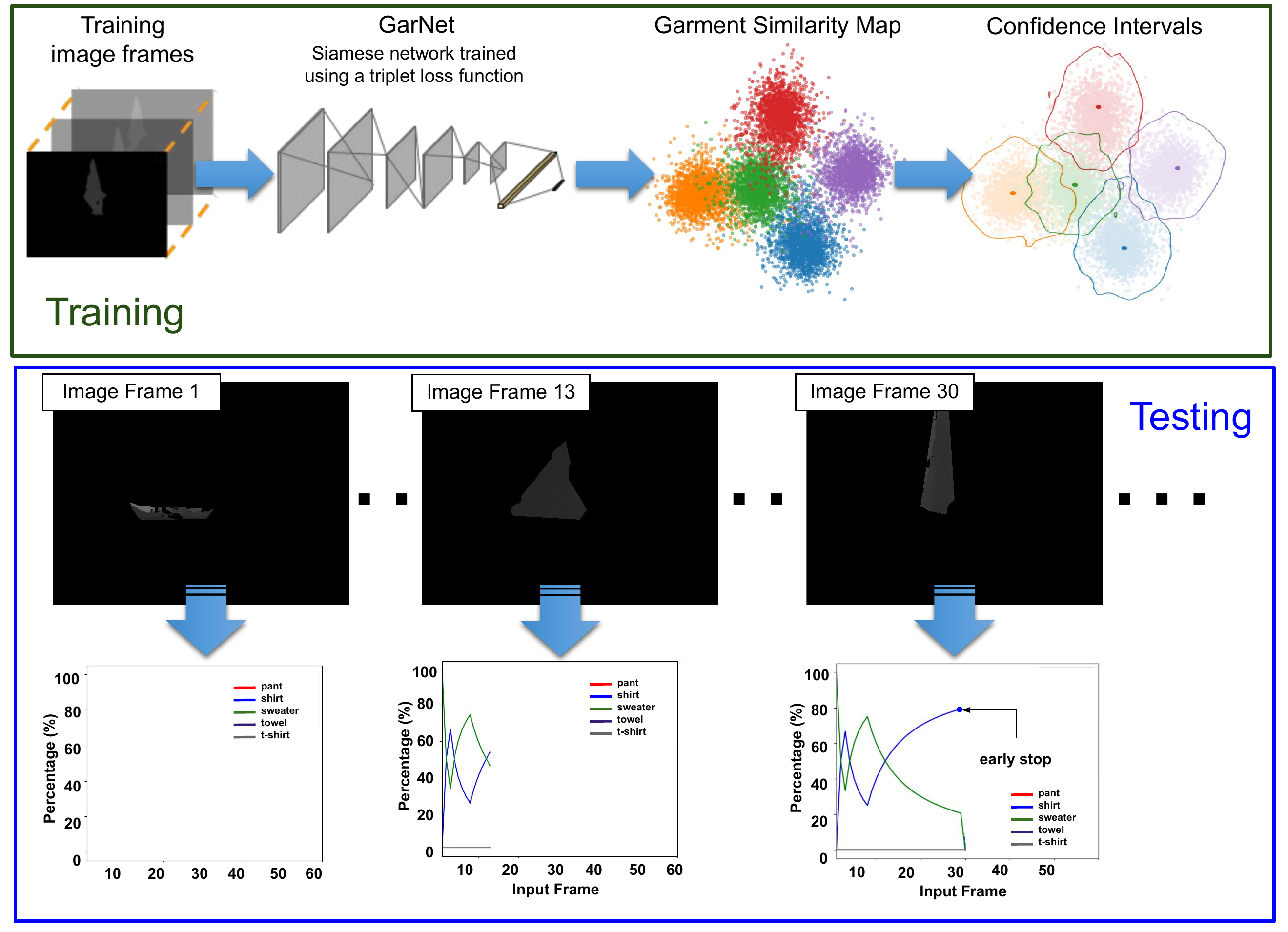}
    \caption{\textit{Top: Training GarNet.} A positive, negative and anchor image samples from a video sequence of the training dataset are input into GarNet. Training, therefore, consists of identifying whether any two of the input triplet come from the same shape or discretised weight categories. GarNet maps input image frames into a Garment Similarity Map (GSM) in which input frames are mapped into clusters if they are similar; otherwise, new points are pulled apart from the cluster. Confidence intervals are computed for each cluster in the GSM as described in Section \ref{sec:conf_cir}. \textit{Bottom: Continuous Perception (Testing GarNet).} An image from a video sequence of the testing dataset is input into a trained GarNet to get the mapping onto the garment similarity map. A video sequence of a garment in our database contains 60 frames. The plots show that GarNet gains confidence in predicting that the tested garment is a shirt. That is, as knowledge is being accumulated into the GSM, most of the decision points belong to the shirt category. In the example shown, the prediction is stopped at frame 30 because 80\% of all decision points belong to the shirt category.}
    \label{fig:pipeline}
\end{figure*}

To test the above hypothesis, we have built a database that consists of RGBD video sequences of a robot grasping and dropping garments on a table\footnote{This database can be downloaded from \url{https://liduanatglasgow.github.io/GarNet/}}, see Fig. \ref{fig:garment_manipulation}. This database simulates a sorting scenario (e.g. \cite{sun2017singleshot, sun2016recognising}) where a robot can sort based on shape or weights. We then train GarNet to learn garments' geometric and physical similarities based on their shape and discretised weight labels. GarNet's objective is thus to cluster garments of the same categories (shapes or discretised weights) together and pull garments of different categories apart using a triplet loss function, and these clusters are mapped into a Garment Similarity Map (GSM). To predict unseen garment shapes and weights, we introduce the concept of decision points which depend on previously mapped points in the similarity map. We use those decision points to devise an early-stop strategy by fitting confidence intervals for each cluster and allow us to determine whether decision points are within a statistical significant interval around a given cluster. An overview of our approach is shown in Figure~\ref{fig:pipeline}.

The contributions of this paper are threefold;

\begin{enumerate}
    \item we have advanced the state-of-art by introducing GarNet that improves the prediction accuracy from 70.8\% to 92\% for shape classification;
    \item GarNet can visually estimate weights of unseen garments with a 95.5\% prediction accuracy;
    \item we propose an early stop strategy so that GarNet is faster during inference compared to the state-of-art, where a frame can be processed in 0.1 seconds.
\end{enumerate}

\section{Related Work}

Previous approaches have proposed learning geometric and physical properties of deformable objects before manipulation. Geometric characteristics of garments include shapes\cite{sun2017singleshot}\cite{MARTINEZ2019220}\cite{sun2016recognising}, their accessories (e.g. zips, buttons), geometric properties (lengths and heights), and physical characteristics such as weight \cite{runia2020cloth}, stiffness (e.g. bending, stretching and shearing) \cite{wang2011data}, damping factors and elasticity \cite{Tawbe2017AcquisitionAN}. For instance, Mariolis \textit{et al.} \cite{mariolis2015pose} proposed to use a CNN network to classify garment shapes, which are rotated by a dual-arm robot. The CNN network learns garment dynamics via depth images and achieves an accuracy rate of 89\% after training. However, they train and test the network on synthetic datasets of simulated garments and their validation image set contains the garments that are already in the training dataset, failing to generalise for unseen garments of similar classes.

Sun \textit{et al.}\cite{sun2017singleshot} presented an approach where local (local B-spline path and locality-constrained linear coding) and global (shape index, local binary patterns, and topology spatial distances) features are extracted from single-shot images and are used to predict unseen garment shapes. This approach makes use of local and global visual characteristics of garments, such as wrinkle features. Compared to \cite{mariolis2015pose}, their approach does not require interactions with garments, so it is faster to predict shapes and is robust while being presented with unseen samples. However, prediction accuracies are constrained by the inability of the robot to interact with the garments, and no new knowledge can be captured. For this, Sun \textit{et al.} \cite{sun2016recognising} proposed a Gaussian process regression classifier to predict unseen garment shapes. That is, the robot in their experiment shakes or flips and then drops garments on a table to obtain a new state to improve the classification score. If the classification score of a garment is above a threshold, the garment is sorted based on their shape. This approach, therefore, demonstrated that interacting with garments enables an autonomous system to improve its prediction confidence over interactions and leads to higher classification accuracies.

However, \cite{sun2016recognising} captured the garment's state while being static and on a table which results in making the system slow at predicting shapes as it requires multiple interactions. Martinez \textit{et al.}\cite{MARTINEZ2019220} removed this limitation by introducing the concept of continuous perception to enable a robot to predict shapes by continuously observing video frames from an Xtion depth-sensing camera rather than single image frames. They showed higher accuracy in predicting unseen garment shapes compared to \cite{sun2016recognising} and \cite{sun2017singleshot}. However, the limitation in \cite{MARTINEZ2019220} is that they let the robot to observe the entire video sequence before a decision can be made, which means that the robot takes a significant amount of time to predict a garment shape category, and this is given by the length of the video. In their work, they sample a garment for approximately 6 seconds which consists of sampling the garment from a crumpled to a hanging states. In this paper, we, therefore, explore the possibility of adopting the continuous perception paradigm to allow a robot to change its manipulation strategy on the fly, i.e as soon as the class of the garment is predicted even if the garment is still being manipulated from a crumpled to a hanging state.

Simulated environments \cite{runia2020cloth}\cite{hoque2021visuospatial} that model deformable objects, have been used in the literature to learn the physical properties of these objects and extrapolate the learned knowledge into the real world. For example, Ruina \textit{et al.}\cite{runia2020cloth} devised an approach where they predicted the area weights of fabrics by learning the physical similarities between simulated fabrics and real fabrics using a spectral decomposition network (SDN). Closing the gap between a simulation and the real world is effective because the physical property parameters of simulated objects are easily accessible compared to those of real objects. However, their approach is not applicable for online evaluation because it requires aligning the simulation with reality, creating an extra overhead before a prediction can be carried out. Hoque \textit{et al.} \cite{hoque2021visuospatial} propose learning dynamic physical properties of towels via a VisionSpatial Foresight network (VSF), which is trained on simulated towels but tested on real towels. The VSF predicts a sequence of towel deformations and corresponding robot actions based on the towels’ initial and desired configurations. They used  RGB plus depth images instead of RGB images in their experiment, and they obtained a success rate  of $90\%$ on manipulating (flattening) the towels. Even though the robot used to deploy VSF takes unnecessary actions while folding a towel, their proposed approach demonstrates that prior knowledge on understanding geometric and physical properties of deformable objects enables an effective manipulation of those objects. Therefore, This paper investigates whether a continuous perception approach coupled with an early-stop strategy can extend beyond simple fabrics and one single shape class. That is, we propose learning the similarity between garments to predict unseen garment shapes and discretised weights based on a ‘garment similarity map’. Compared to previous works (e.g. \cite{MARTINEZ2019220}, and \cite{sun2016recognising}), our work features an early-stop strategy, where a prediction can be halted earlier without observing the entire interaction that can enable an online manipulation strategy.

\section{GarNet: Garment Similarity Network \label{sec:GarNet} \label{sec:GarNet_main}}

Our proposed Garment similarity Network (GarNet) consists of a Siamese network \cite{koch2015siamese} which clusters garments into groups according to their shape and discretised weight categories. The objective of clustering garments is to learn common geometric and physical features of garments of the same categories. Our GarNet network comprises a residual convolutional block that extracts features from input data and a fully connected layer that maps features onto a 2D \textit{Garment Similarity Map} (GSM). A garment similarity map is a 2D manifold that encodes a garment's geometric and physical characteristics according to its shape and discretised weight categories. Garments of the same categories are clustered together, while garments of different categories are pulled apart. Figure \ref{fig:train_cluster} shows the garment similarity map in our experiments where each cluster in the map is called a Garment Cluster ($\mathcal{GC}$). Our GarNet training process is expressed mathematically as:

\begin{equation}
    \mathcal{P}=\mathcal{F}[\mathcal{C}(I)]
    \label{eq:garnet_and_gsp}
\end{equation}
where $\mathcal{C}$ denotes residual convolutional layers, $\mathcal{F}$, fully connected layers, and $I$, input video frames. We define $\mathcal{P}$ as a garment similarity point ($\mathcal{GSP}$). Each frame in the input video sequence of the garments is converted into one garment similarity point ($\mathcal{GSP}$). We also define a Garment Similarity Distance ($\mathcal{GSD}$) as:

\begin{equation}
    \mathcal{GSD}(x,y)=\mathcal{P}_i-\mathcal{P}_j
    \label{eq:gsd}
\end{equation}
where $i$ and $j$ are the $ith$ and $jth$ garment similarity point. $\mathcal{GSD}$ increases between garments with different labels and decreases between garments with the same labels. Therefore, to train GarNet, we use a triplet loss \cite{hoffer2015deep}:

\begin{equation}
\begin{aligned}
    {PP} &=\abs{\mathcal{P}_{positive}-\mathcal{P}_{anchor}} \\
    {NP} &=\abs{\mathcal{P}_{negative}-\mathcal{P}_{anchor}} \\
    {TripletLoss} &=max (0, PP-NP+margin)\label{eq:loss}
\end{aligned}
\end{equation}
where $PP$ is a positive pair between positive and anchor samples and $NP$ is a negative pair between negative and anchor samples. A positive sample is an image of a garment of the same category with respect to the anchor. A negative sample is an image of a garment with different category with respect to the anchor. The $margin$ is a value that promotes the network to learn to map positive and negative samples further away from each other. In this experiment, we set this $margin$ to 1 as suggested in \cite{runia2020cloth} experiments.

Two GarNets are trained separately and independently predict the shapes and discretised weights independently. That is, the GarNet for shape predictions is trained on the shape categories, while GarNet for discretised weight predictions is trained on discretised weight categories, e.g. light, medium and heavy weights.

\subsection{Garment Cluster Confidence Intervals \label{sec:conf_cir}}

To decide which category (either shapes or discretised weights) a mapped garment similarity point in the similarity map belongs to, we propose to fit statistical confidence intervals to each garment cluster in this map. That is, we define a confidence interval using a non-parametric probability density function for each garment cluster, $\mathcal{GC}$ via a kernel density estimator \cite{10.1214/aoms/1177728190} that is defined as:

\begin{equation}
\begin{aligned}
    \hat{f_h}(\mathcal{GC}) &=\frac{1}{n}\sum_{i=1}^{n}K_h(\mathcal{P}-\mathcal{P}_i) \\
     &=\frac{1}{nk}\sum_{i=1}^{n}K(\frac{\mathcal{P}-\mathcal{P}_i}{h})
\label{eq:conf_circle}
\end{aligned}
\end{equation}
where $\mathcal{GC}$ is the garment cluster, $K$ is a Gaussian kernel, $h>0$ is a smoothing parameter called bandwidth which regulates the amplitude of confidence intervals, and $\hat{f_h}$ is an estimated probability density function for a garment cluster. We have conducted an ablation study on the confidence interval's bandwidths ($h$) and results are presented in section \ref{sec:confi_circle_ablation}. After training a GarNet, the centroid of each garment cluster is defined as:

\begin{equation}
    \mathcal{GC}_{mean}=(\frac{1}{m}\sum_{i=1}^m x_{\mathcal{P}_i}, \frac{1}{m}\sum_{i=1}^m y_{\mathcal{P}_i})
\end{equation}
where $\mathcal{GC}_{mean}$ is the mean value of garment similarity points mapped from one garment cluster (in Figure \ref{fig:train_cluster}) and $m$ is the number of garment similarity points in the cluster.

In our experiments, we directly input unseen garments image frames acquired by the robot to GarNet to map them into the garment similarity map. To decide the shapes and discretised weights, we define a Decision Point ($\mathcal{DP}$) that is the mean value of garment similarity points ($\mathcal{GSP}$s):

\begin{equation}
    \mathcal{DP}=(\frac{1}{n}\sum_{i=1}^n x_{\mathcal{P}_i}, \frac{1}{n}\sum_{i=1}^n y_{\mathcal{P}_i})
    \label{eq:decision_point}
\end{equation}
where $n$ is the total number of frames observed. To predict the shapes and discretised weights, we find whether a $\mathcal{DP}$ is within any confidence interval and has the minimum distance to the confidence interval's $\mathcal{GC}_{mean}$. For this paper, we use the Euclidean distance to evaluate how close a $\mathcal{DP}$ is with respect to $\mathcal{GC}_{mean}$.

Each video sequence has 60 frames; therefore, we will have 60 decision points. To predict the shape and discretised weight, we establish that a predicted category should have at least $80\%$ of decision points belonging to a garment cluster. If none of the categories fulfils this requirement, we denote that the observed garment does not have a known class. That is, if a decision point is outside any confidence interval, the network is not confident about which category the input garment belongs to. By clustering garments and defining confidence intervals, it is possible to define an early-stop strategy to allow a robotic system to stop its execution if it is confident about the garment shape or weight. After observing a number of image frames of a garment, if any of the trained categories takes $80\%$ of the decision points, the process is terminated, and the category is chosen as the predicted category.

\section{Experiments}

\subsection{GarNet Architecture}
We implement our code in Pytorch \cite{NEURIPS2019_9015}.  Our GarNet comprises a ResNet18 \cite{he2015deep} as a feature extraction and fully connected networks (FC). The FC networks comprise three linear layers, where a PReLU activation layer is placed between adjacent linear layers. The source code for GarNet and experimental scripts are available at \url{https://liduanatglasgow.github.io/GarNet/}.

We use an Intel-i7 equipped computer with an Nvidia GTX 1080 Ti to train the network. We use the Adam optimizer with an initial learning rate of $\num{e-3}$, controlled by a learning scheduler with a decay rate of $\num{e-1}$ and a step size of 8. The network is trained for 270K iterations, taking approximately 30 minutes.

\subsection{Data Collection and Experiments \label{sec:experiments}}

\begin{figure*}
    \centering
    \includegraphics[width=0.9\textwidth]{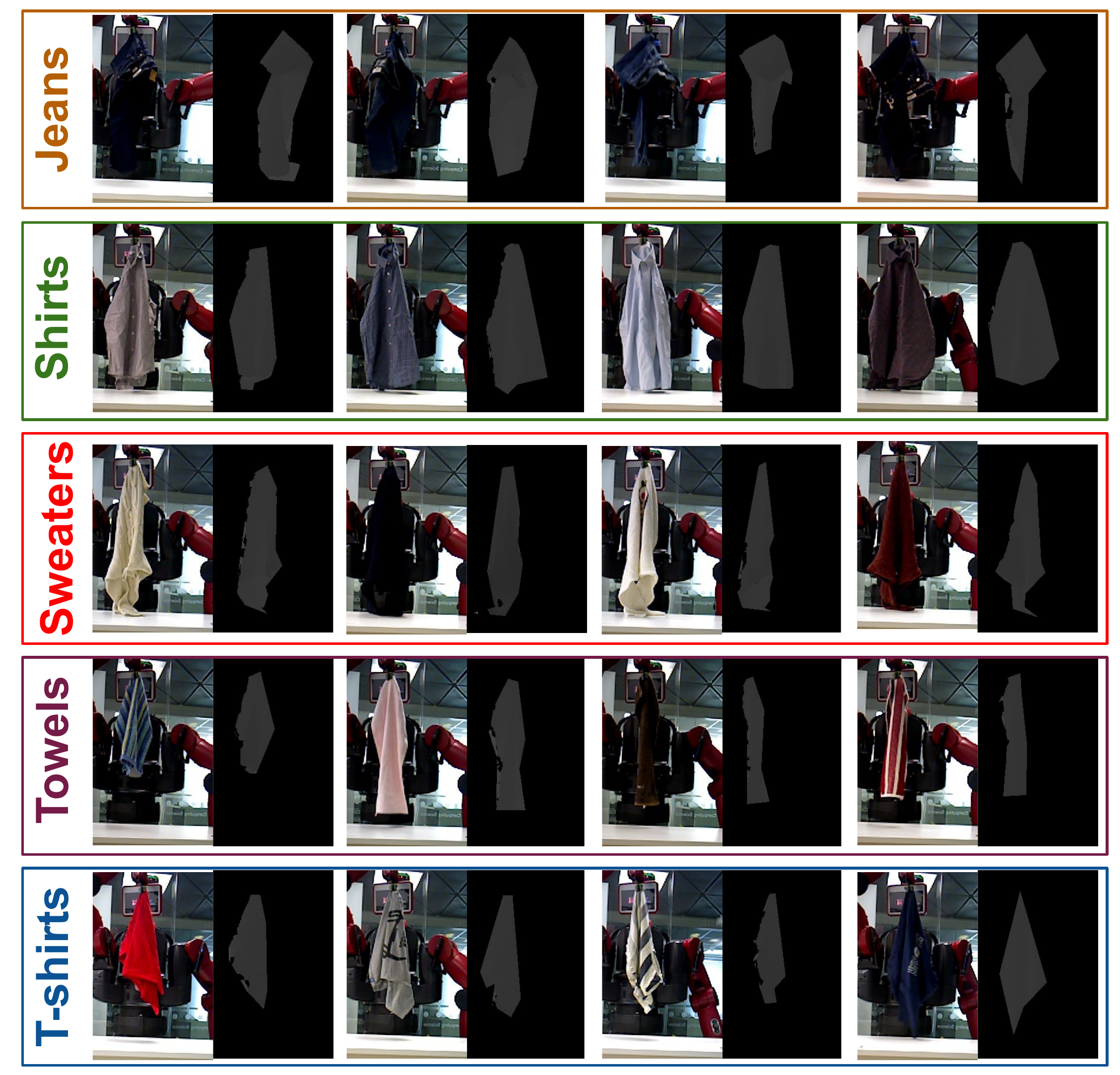}
    \caption{\textit{The garment database used in our experiments}: Five categories (pants (jeans), shirts, sweaters, towels and t-shirts), and each category has four garment instances. For each garment instance, we show an RGB image frame and its corresponding segmented depth image}
    \label{fig:data_fig}
\end{figure*}

The video database in this experiment consists of 20 garments of five different shapes, namely, pants, shirts, sweaters, towels and t-shirts. Figure \ref{fig:data_fig} shows garment samples for each garment instance in our database, i.e. five categories and four garments for each category. For each shape, there are four garments of different colours and materials. We used an electric scale to weigh every garment and divide their weights into three discretised levels, namely, light, medium and heavy weights. Therefore, in these experiments, we do not predict the real weight values of tested garments but predict the discretised weights levels in order to enable a robot sort garments as we do before putting garments into a washing machine.

To validate our network and test our hypothesis (Section \ref{sec:intro}), we propose to carry out a leave-one-out cross-validation methodology. That is, we group all garments into four groups and each garment category as shown in Fig. \ref{fig:data_fig}, has four different garment instances. Hence, four experiments are conducted, where three groups served as training groups and one group served as a testing group. We averaged accuracies for each category output from the four experiments and used testing group to validate the classification performance of our approach. For each of the four experiments, the training group represent 80\% of our video sequence database, while the testing group represent 20\% of our video sequence database. 

We have used a Baxter robot to manipulate garments. Garments are initially placed on a table, and, next, they are lifted to $1m$ above the table and dropped from this height. An Xtion depth-sensing camera is used to capture garment video sequences. Each garment is captured ten times, which means that the grasp-and-drop operation is conducted ten times. There are 200 videos in total, and each video contains 60 frames (sampling frequency is 10Hz; video sequence length is 6 seconds). Therefore there are 12,000 image frames in total. Figure \ref{fig:garment_manipulation} shows the experimental setup of the robot grasping and dropping garments.

Our experiments include 50 unseen garment videos containing ten videos for each of the five categories for each of the four leave-one-out cross-validation experiments. For each video sequence, we predict the shape and discretised weight of the garment in the video. Therefore, we have ten predictions for each category (one prediction for each video) and 50 predictions in total. The prediction accuracy for each category is defined as the percentage of correctly predicted videos of each category.

\section{Results}

\begin{figure*}[t]
    \centering
    \includegraphics[width=0.6\textwidth,keepaspectratio]{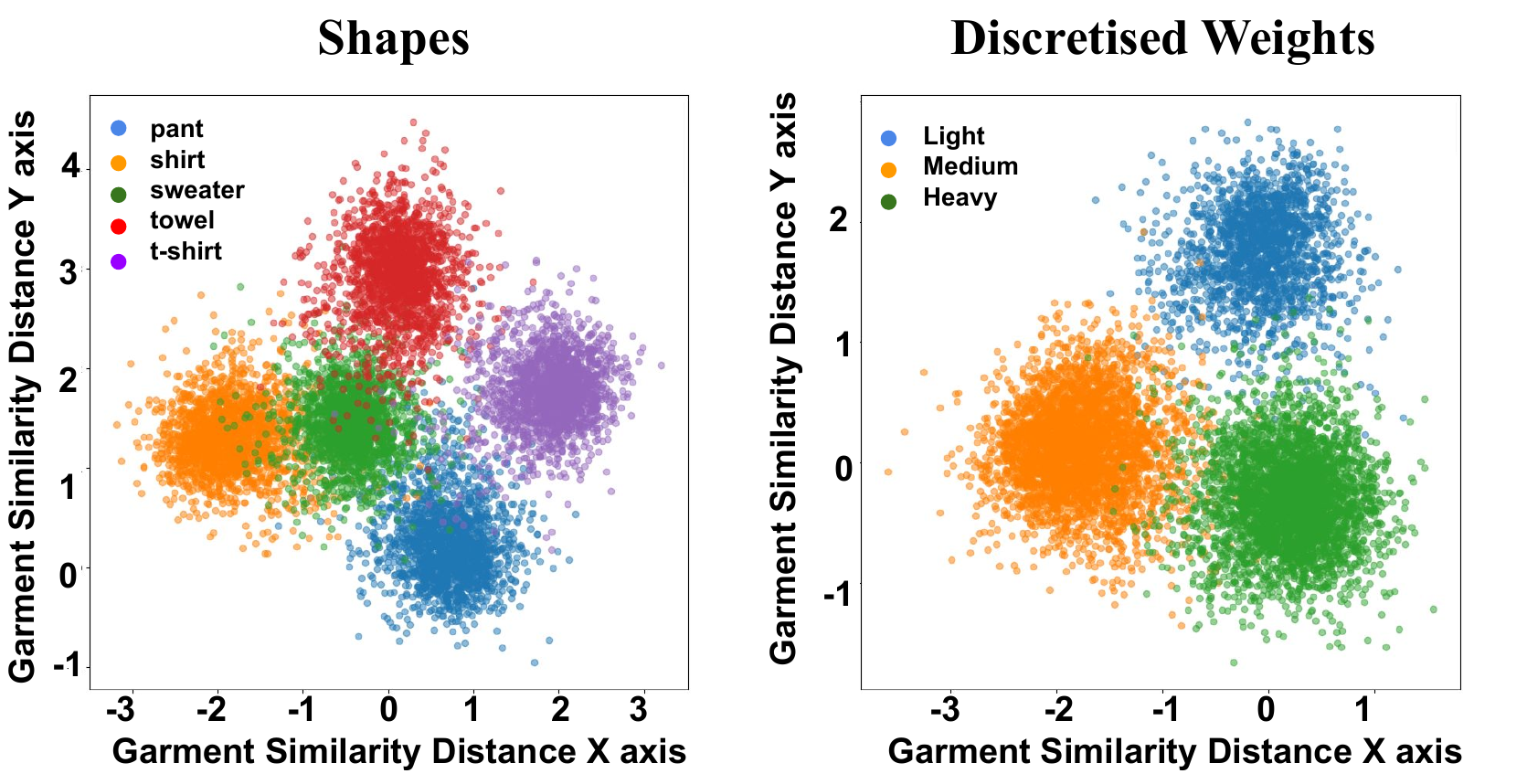}
    \caption{The Garment Similarity Map (GSM) after training GarNet.}
    \label{fig:train_cluster}
\end{figure*}

\subsection{GarNet Training}
While training GarNet, our approach achieves a validation classification performance of $93.9\%$ for shapes and $94.9\%$ for discretised weights. Figure \ref{fig:train_cluster} shows the mappings of testing garments onto the similarity map, where it is possible to observe that garments of different categories are pulled apart; while garments of the same categories are clustered together. These results confirm that GarNet coupled with a triplet loss function (Eq. \ref{eq:loss}) is able to extract physical dissimilarities between categories while maintaining inter-class physical properties within well-defined clusters. 

\subsection{Continuous Perception Experiments}

\begin{figure*}[t]
    \centering
    \includegraphics[width=0.7\textwidth]{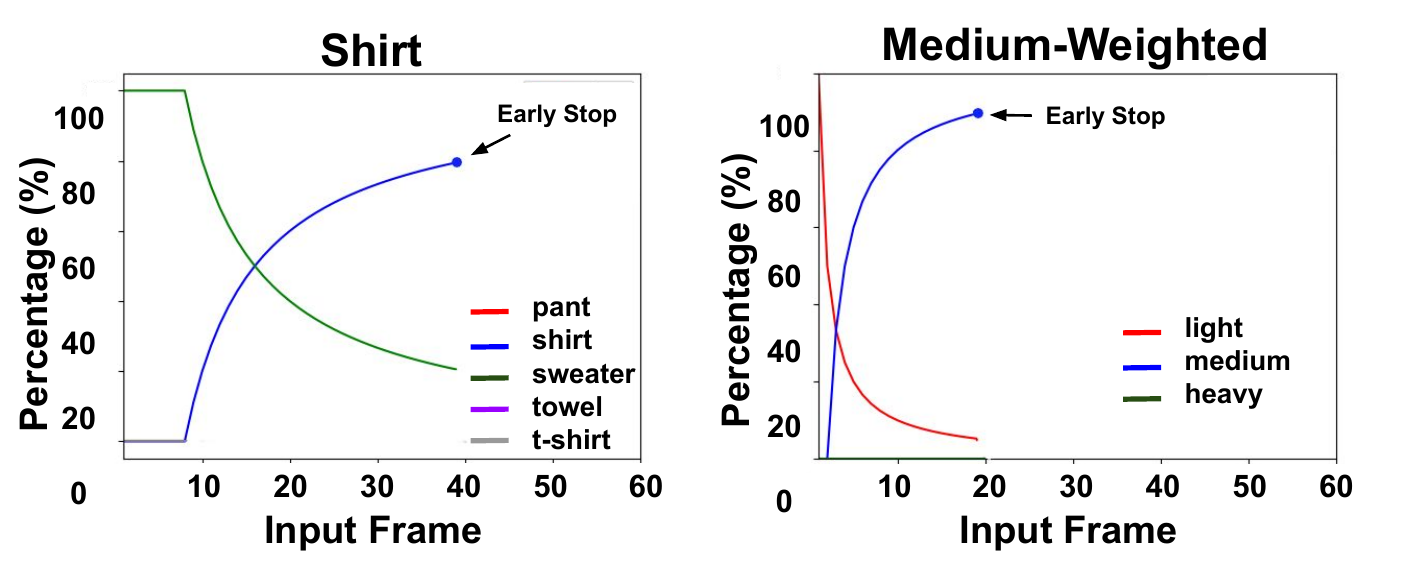}
    \caption{Examples of the early-stop strategy proposed in this paper. The early-stop strategy allows the robotic system to make a prediction without observing the full motion. As observed in both plots, GarNet becomes confident over time, and the early-stop strategy activates if 80\% of decision points in the garment similarity map are within a correct category. Plots for all unseen predictions can be accessed at \url{https://liduanatglasgow.github.io/GarNet/}}
    \label{fig:contin_percep}
\end{figure*}

Examples of our experimental results are shown in Figure \ref{fig:contin_percep}. We can see that although GarNet does not recognise garments correctly from the beginning (because most of the decision points belong to an incorrect category), GarNet gradually gains confidence in predicting a correct category for each garment because more decision points are within the correct category and eventually percentages of decision points are over 80\%. The classification task is consequently stopped early, and the system does not need to observe the full robot motion (and video sequence) to make a correct prediction of the garment class.

We conduct two ablation studies for the continuous perception experiment. The first study is about comparing predictions only on local garment similarity points ($\mathcal{GSP}$s) rather than on decision points ($\mathcal{DP}$s). The second ablation study compares the performance of GarNet trained on RGB and depth images.

Tables \ref{tab:shape} and \ref{tab:weight} show the results of the leave-one-out cross-validation experiments, where the network achieved $92\%$ for shape classification and $96\%$ for discretised weight classification. The results show that our network has an expected ability to classify shapes and discretised weights of unseen garments.

We use ‘decision points’, $\mathcal{DP}$s, (Eq. \ref{eq:decision_point}) to make predictions on unseen garment shapes and discretised weights. That is, the position of a decision point on the garment similarity map (in Figure \ref{fig:train_cluster}) depends on all previously observed image frames rather than on currently observed image frames. From Tables \ref{tab:shape} and \ref{tab:weight}, we can observe that using decision points has better performance than using garment similarity points ($92\%$ vs $78\%$ for shapes and $95.5\%$ vs. $80\%$ for discretised weights, respectively). This shows that GarNet benefits from using accumulated knowledge via decision points rather than local and episodic knowledge as in \cite{mariolis2015pose, sun2017singleshot}.

\begin{table}[t]
    \centering
    \caption{Table: Prediction Results (shapes)}
    \begin{tabular}{|c|c|c|c|c|c|}
    \hline
        Category &  depth, $\mathcal{DP}$& depth, $\mathcal{GSP}$& RGB, $\mathcal{DP}$ & RGB, $\mathcal{GSP}$\\
    \hline
        \textit{pants} & {97.5\%}&  {82.5\%} &  {97.5\%} &  {87.5\%}\\
    \hline
        \textit{shirts} &  {72.5\%} &  {75\%} &  {87.5\%}&  {62.5\%}\\
    \hline
        \textit{sweaters}&  {97.5\%} &  {85\%} &  {25\%} &  {20\%}\\
    \hline
        \textit{towels} &  {92.5\%} &   {65\%} &  {50\%}&  {17.5\%}\\
    \hline
        \textit{t-shirts} &  {100\%} &   {82.5\%} &  {32.5\%}&  {22.5\%}\\
    \hline
    \textbf{Average} &  {\textbf{92\%}} &  {78\%} &  {53.5\%} &  {42\%}\\
    \hline 
    \end{tabular}
    \label{tab:shape}
\end{table}

\begin{table}[t]
    \centering
    \caption{Table: Prediction Results (discretised weights)}
    \begin{tabular}{|c|c|c|c|c|c|}
    \hline
        Category &  depth, $\mathcal{DP}$& depth, $\mathcal{GSP}$& RGB, $\mathcal{DP}$& RGB, $\mathcal{GSP}$\\ 
    \hline
        \textit{lights}&  {97.5\%}&  {50\%}&  {37.5\%}&  {5\%}\\
    \hline
        \textit{mediums}&  {93.75\%}&  {87.5\%}& {72.5\%}&  {58.75\%} \\
    \hline
        \textit{heavies}&  {96.25\%}&  {87.5\%}&  {71.25\%}&  {71.25\%}\\
    \hline
    \textbf{Average}&     {\textbf{95.5\%}}&  {80\%} &  {65\%}&  {53\%}\\
    \hline 
    \end{tabular}
    \label{tab:weight}
\end{table}

To investigate whether the type of image affects the overall prediction of a garment class, we trained GarNet using RGB and depth images. Tables \ref{tab:shape} and \ref{tab:weight} show that a GarNet trained on depth images outperforms a GarNet trained on RGB images (92\% vs 53.5\% for shapes and 95.5\% vs 65\% for discretised weights, respectively). The increase in performance is because depth images capture structural and dynamic information of the garment being manipulated and are better suited to capture the physical properties of garments as opposed to RGB images as proposed by \cite{runia2020cloth}.

Note that the inter-class variability for the pants category is consistent (i.e. we use jeans for this category, see Fig. \ref{fig:data_fig}, top row). Therefore, classification scores in Table \ref{tab:shape} for pants are high across the ablation studies with respect to other shape categories of which they have high inter-class variability. This result shows that in order to generalise to unseen garments, depth images and decision points offer the best combination for the continuous perception task.

\subsection{Ablation study on the confidence intervals bandwidths \label{sec:confi_circle_ablation}}

A bandwidth, as defined in section \ref{sec:conf_cir}, determines the size of a confidence interval. In this study, we, therefore, evaluate the effect of the bandwidth selection with respect to the performance of GarNet. A confidence interval of a garment cluster is a region in the garment similarity map of which a certain percentage of $\mathcal{GSP}$s are grouped together. 

A decrease in the bandwidth value denotes a decrease in the percentage of $\mathcal{GSP}$s included within the garment cluster. An increase in the bandwidth means that almost all $\mathcal{GSP}$s should be included, while a small portion of points are relatively far away from a garment cluster. This means that a confidence interval may overlap with other confidence intervals, or even multiple confidence intervals will be generated for one garment cluster. In this case, the performance of GarNet will have an impact in the final classification prediction.

\begin{figure*}[t]
    \centering
    \includegraphics[width=0.65\textwidth]{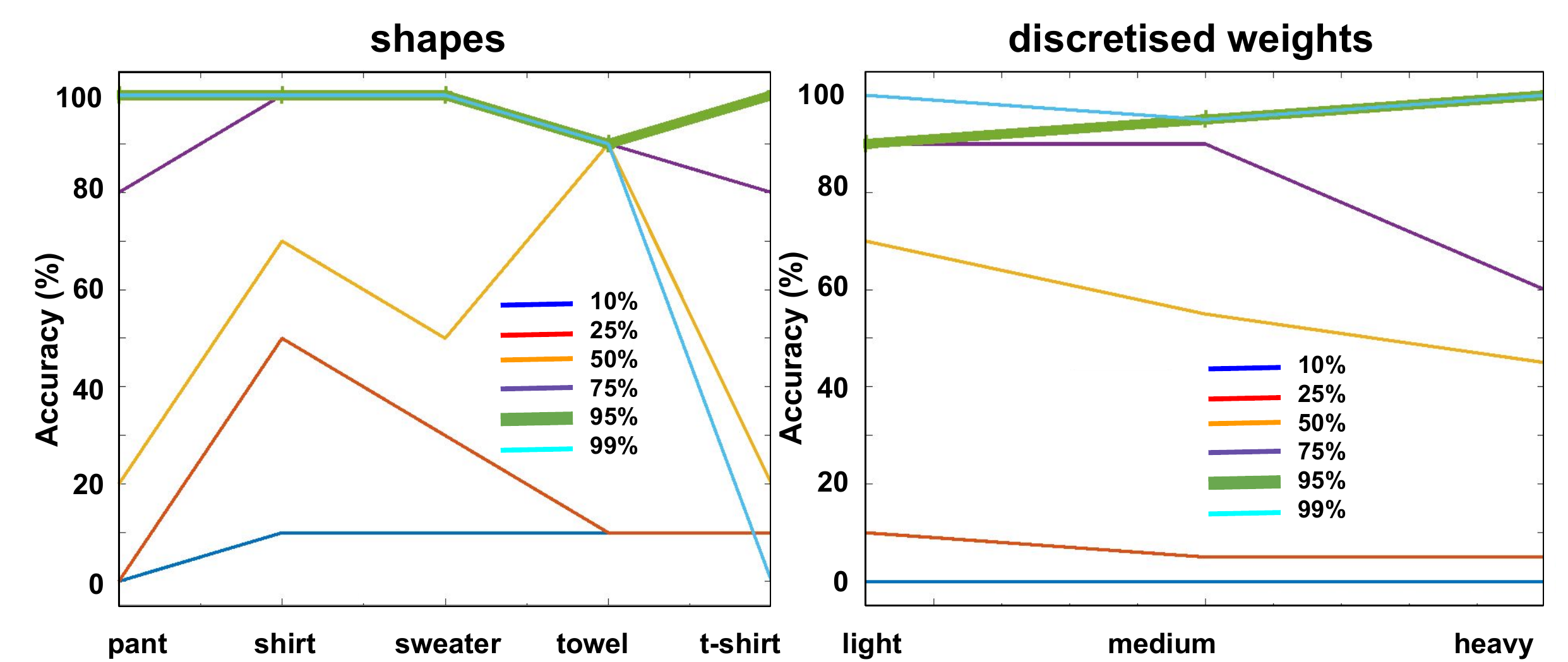}
    \caption{Bandwidth Ablation Study. \textit{Left}: Bandwidth values versus accuracy for shape classification. \textit{Right}: Bandwidth values versus Accuracy for discretised weights classification.}
    \label{fig:bandwidth_study}
\end{figure*}

In Figure \ref{fig:bandwidth_study}, we find that a bandwidth of $95\%$ has the best performance, and we use $95\%$ as the bandwidth for the rest of the experiments. However, at a $99\%$ bandwidth, we can observe that the prediction accuracy drops while classifying shapes; this is because $\mathcal{GSP}$s are grouped into incorrect categories.

\subsection{Comparison with the state-of-art}
We have compared our results with the results from \cite{sun2017singleshot}\cite{MARTINEZ2019220}\cite{sun2016recognising}\cite{visapp22}. As observed in Table \ref{tab:soa}, our GarNet outperforms previous work on predicting unseen garment shapes. There are several reasons why our network has the best performance and we discuss these below.

\subsubsection{The use of a garment similarity map to encode knowledge of garment shapes and weights} Inspired by \cite{runia2020cloth}, where the authors proposed learning the physical similarity between simulated fabrics and predicting physical properties of real fabrics, we find that the similarity network effectively predicts unseen deformable objects, such as garments and fabrics. In our previous work \cite{visapp22}, where we  focused on utilising solely classification approach rather than the clustering approach in our network, we found that our clustering approach presented in this paper improves over the classification approach. Compared with a traditional classifier that regresses embeddings of data into labels (which is equivalent to asking which shape/discretised weight classes the data belongs to), our GarNet network learns geometric and physical characteristics that makes them the same or different (which is equivalent to asking why the data presents the same or different shapes/discretised weights). Therefore, for unseen garments, the network only needs to decide similarities of the garments for each garment cluster rather than classify them into certain classes, reducing the prediction difficulty.

\subsubsection{Continuous Perception.} Traditional methods such as \cite{sun2017singleshot,mariolis2015pose} that predict shapes and weights of garments are based on static garment features such as wrinkles, outlines, creases, to name a few. Instead, we propose to carry out predictions on encoded knowledge in the GSM  while learning the dynamic properties of garments.

\subsubsection{Early-Stop strategy.} Compared with \cite{sun2016recognising}\cite{MARTINEZ2019220}, where the proposed approaches consist of having a robot observing the entire interaction with a garment, our approach only requires a robot to observe interactions partly if termination requirements are satisfied. Therefore, our approach has a mechanism to stop a manipulation on the fly as GarNet can process images every 0.1 seconds, taking an average of 4 seconds to generate a prediction.

\begin{table}[!t]
    \centering
    \caption{Comparisons with the state-of-art.}
    \begin{tabular}{|c|c|}
    \hline
    Method & Accuracy (\%) \\ \hline
       CNN-LSTM (classification) \cite{visapp22} & $48\%$\\ \hline
       Interactive Perception\cite{sun2016recognising}  & $64.2\%$ \\ \hline
       Single-shot category  recognition\cite{sun2017singleshot} & $67.0\%$ \\ \hline
       Continuous Perception in \cite{MARTINEZ2019220} & $70.8\%$ \\ \hline
       \textbf{GarNet (Continuous Perception, Ours)} & \textbf{$92.0\%$} \\ \hline
    \end{tabular}
    \label{tab:soa}
\end{table}
 
\section{Conclusion}
We have presented a garment similarity network (GarNet) that learns the similarity of the garments and predicts continuously their shapes and their visually perceived weights. We have also introduced a Garment Similarity Map ($\mathcal{GSM}$) that encodes garment shapes and weights knowledge into clusters. These clusters were then used to decide on which cluster unseen garment samples belongs to heuristically. Our experimental validation shows that, GarNet obtains high prediction accuracies while classifying shapes (92\%) and discretised weights (95.5\%), Fig. \ref{fig:contin_percep}. Similarly, we have compared GarNet's performance with the state of the art, and GarNet showed an increase of 21.2\% of classification accuracy performance (Table \ref{tab:soa}).

Compared with previous work on continuous perception \cite{MARTINEZ2019220}, GarNet has the advantage of an ‘early stop’ strategy. That is, a robot does not need to observe the full motion (video sequences) to make predictions which could enable robots to be more responsive and effective while manipulating garments and deformable objects in a laundry pipeline. However, GarNet, in this paper, does not support online learning of unknown garment shapes. For instance, we train GarNet on five shape categories, and it can predict shapes of unseen garments from those categories.

In future work, we plan to devise an online-learning approach for GarNet to investigate complex manipulations interactions (enabled by a behaviour-based reinforcement learning agent \cite{pore2020simple}), such as twisting garments, shaking garments or rotating garments. From those interactions, differences in stretching and bending characteristics of garments can be exploited to evaluate garments’ stiffness parameters, which can potentially help to develop a robot dexterous garment manipulation approach for folding \cite{doumanoglou2016folding} \cite{6942541}, flattening \cite{7138998}, to name a few, that requires fewer iterations. Furthermore, knowledge of the garments weights, i.e. whether it is light, medium or heavy, can enable a robot to plan for these complex manipulation interactions since it will be possible to reduce the search space while estimating the dynamic physical properties of garments.

\section{Acknowledgement}
We would like to thank George Killick and Nikos Pitsillos for their valuable comments while reviewing this paper.

\bibliographystyle{IEEEtran}
\bibliography{references.bib}
\end{document}